# Data Separability for Neural Network Classifiers and the Development of a Separability Index

Shuyue Guan, Murray Loew*, *Fellow, IEEE*, and Hanseok Ko

**Abstract**—In machine learning, the performance of a classifier depends on both the classifier model and the dataset. For a specific neural network classifier, the training process varies with the training set used; some training data make training accuracy fast converged to high values, while some data may lead to slowly converged to lower accuracy. To quantify this phenomenon, we created the Distance-based Separability Index (DSI), which is independent of the classifier model, to measure the separability of datasets. In this paper, we consider the situation where different classes of data are mixed together in the same distribution is most difficult for classifiers to separate, and we show that the DSI can indicate whether data belonging to different classes have similar distributions. When comparing our proposed approach with several existing separability/complexity measures using synthetic and real datasets, the results show the DSI is an effective separability measure. We also discussed possible applications of the DSI in the fields of data science, machine learning, and deep learning.

————————————— ◆ —————————————

## 1 INTRODUCTION

DATA and models are the two main foundations of machine learning and deep learning. Models learn knowledges (patterns) from datasets. An example of this is a convolutional neural network (CNN) classifier, which learns how to recognize images from different classes. There are two respects to examine the learning process: capability of the classifier and the separability of dataset. Separability is an intrinsic characteristic of a dataset to describe how data points belonging to different classes mix with each other. The learning outcomes are highly relied on the two respects. For a specific model, the learning capability is fixed so that the training process depends on the training data. As reported by Zhang et al. (2016) [1], time to convergence for the same training loss on random labels on the CIFAR-10 dataset was slower than training on true labels. It is not surprising that the performance of a given model varies between different training datasets depending on their separability. For example, in a two-class problem, if the scattering area for each class has no overlap, one straight line can completely separate the data points (Figure 1a). For the distribution shown in Figure 1b, however, a single straight line cannot separate the data points successfully, but a combination of many lines can.

In machine learning and deep learning, it is common for some datasets to be more difficult to train on for a specific classifier. The difficulty of training is shown by the needs of more learning times (*e.g.* epochs for deep learning) to reach the same accuracy (or loss) value or/and, to finally reach a lower accuracy (or higher loss). To quantify train-ing difficulty in this paper, we define training distinctness (TD) as the average training accuracy during the training process; a lower TD means that a dataset is more difficult to train, and this difficulty can be used to reflect the separability of the dataset. Training difficulty, however, also depends on the model employed. Hence, an intrinsic measure of separability that is independent of a classifier model is required. In other words, it is significant to be able to measure the separability of a dataset without using a classifier model.

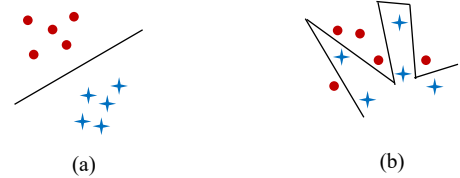

Figure 1 Different separability of two datasets

- *Shuyue Guan and Murray Loew are with the Department of Biomedical Engineering, the George Washington University, Washington DC, USA.*
  *E-mail: frankshuyueguan@gwu.edu, loew@gwu.edu.*
- *Hanseok Ko is with the Department of Electrical and Computer Engineering, Korea University, Anamro 145, Sungbuku, Seoul, Korea.*
  *E-mail: hsko@korea.ac.kr.*

*(\*Corresponding author: Murray Loew.)*

## 2 RELATED WORK

To the best of our knowledge, the number of separability measures for datasets that do not use classifiers is limited and, to date, there have been significantly fewer studies on data separability than on classifiers. The Fisher discriminant ratio (FDR) [2] measures the separability of data using the mean and standard deviation (SD) of each class. It has been used in many studies, but it fails in some cases (*e.g.* Class 1 data points are scattered around Class 2 data points as a circle; their FDR $\approx 0$) because it extracts only two features from the data. A more general issue than that of data separability is data complexity, which measures not only the relationship between classes but also the data distribution in space. Ho and Basu (2002) [3] conducted a ground-breaking review of data complexity measures. They reported measures for classification difficulty, including those associated with the geometrical complexity of class boundaries. Recently, Lorena et al. (2019) [4] summarized existing methods for the measurement of classification complexity. In the survey, most complexity measures have been grouped in six categories: feature-based, linearity, neighborhood, network, dimensionality, and class imbalance measures (TABLE 1). For example, the FDR is a feature-based measure, and the geometric separability index (GSI) [5] proposed by Thornton (1997) is considered a neighbor-based measure. Other ungrouped measures discussed in Lorena's paper have similar characteristics to the grouped measures or may have large time cost. Each of these methods has possible drawbacks. In particular, the features extracted from data for the six categories of feature-based measures may not accurately describe some key characteristics of the data; some linearity measures depend on the classifier used, such as support-vector machines (SVMs); neighborhood measures may only show local information; some network measures may also be affected by local relationships between classes depending on the computational methods employed; dimensionality measures are not strongly related to classification complexity; and, class imbalance measures do not take the distribution of data into account.

In this paper, we create a novel separability measure for multi-class datasets and verify it by comparing with training difficulty and other measures using synthetic and real (CIFAR-10/100) datasets. Since several previous studies [5]–[9] have used the term separability index (SI), we refer to our measure as the distance-based separability index (DSI). The DSI measure is similar in some respects to the network measures because it represents the universal relations between the data points. In Section 4.2, the proposed DSI is used to explain the necessity of image preprocessing in CNN classification. In Section 5.4, we demonstrate that DSI could help break the 'black-box' of neural networks – to understand how data separability changes after passing through each layer of a neural network. In general, data separability/complexity measures have a wide applicability and are not limited to simply understanding the data; for example, they can also be applied to the selection of classifiers [10]–[13] and features for classification [14], [15].

| Category | Name | Code |
|---|---|---|
| Feature-based | Maximum Fisher's discriminant ratio | F1 |
| | Directional vector maximum Fisher's discriminant ratio | F1v |
| | Volume of overlapping region | F2 |
| | Maximum individual feature efficiency | F3 |
| | Collective feature efficiency | F4 |
| Linearity | Sum of the error distance for linear programming | L1 |
| | Error rate of the linear classifier | L2 |
| | Non-linearity of the linear classifier | L3 |
| Neighborhood | Fraction of borderline points | N1 |
| | Ratio of intra/extra class NN distance | N2 |
| | Error rate of the NN classifier | N3 |
| | Non-linearity of the NN classifier | N4 |
| | Fraction of hyperspheres covering the data | T1 |
| | Local set average cardinality | LSC |
| Network | Density | Densit |
| | Clustering coefficient | ClsCo |
| | Hubs | Hubs |
| Dimensionality | Average number of features per dimension | T2 |
| | Average number of PCA dimensions per point | T3 |
| | Ratio of the PCA dimension to the original dimension | T4 |
| Class imbalance | Entropy of class proportions | C1 |
| | Imbalance ratio | C2 |

## 3 METHOD

### 3.1 Methodological development

In a two-class dataset, according to the maximum entropy principle (MEP), the most difficult situation to separate the data is when the two classes of data are scattered and mixed together with the same distribution. In this situation, the proportion of each class in every small region is equal, and the system has maximum entropy. In extreme cases, to obtain 100% classification accuracy, the classifier must separate each data point into individual regions (Figure 2).

Therefore, based on the MEP, we could define data separability as the inverse of a system's entropy. To calculate entropy, the space is divided into many small regions. Then, the proportions of each class in every small region can be considered as their occurrence probabilities. The system's entropy can be derived from those probabilities. In high dimensional space (*e.g.* image data), however, the number of small regions grows exponentially. For example, the space for 32x32 pixels 8-bit RGB images has 3,072 dimensions. If each dimension (ranging from 0 to 255, integer) is divided into 32 intervals, the total number of small regions is $32^{3072} \approx 6.62 \times 10^{4623}$. It is thus impossible to compute the system's entropy in this way.

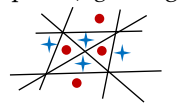

Figure 2 Two-class dataset with maximum entropy

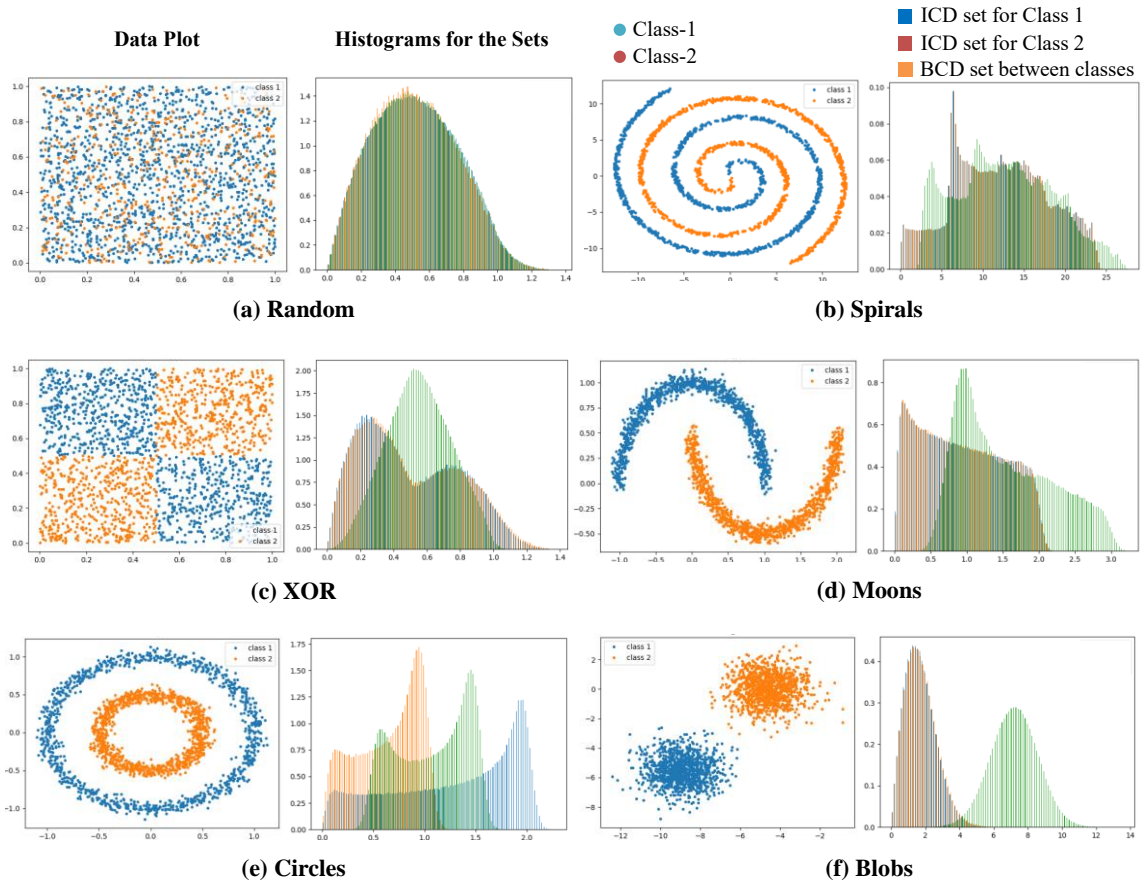

Figure 3 Typical two-class datasets and their ICD and BCD set distributions

Alternatively, we proposed the DSI as a substitute for entropy to analyze how two classes of data are mixed together. The metric for distance is Euclidean ($l^2$-norm). Consider the two classes $X$ and $Y$ with the same distribution (distributions have the same shape, position and covered region, *i.e.* same probability density function) and have sufficient data points. If $X$ and $Y$ have $N_x$ and $N_y$ data points, respectively, we can define the intra-class distance (ICD) set, which is the set of distances between any two points in the same class ($X$), as follows:

$$\{d_x\} = \left\{ \|x_i - x_j\|_2 | x_i, x_j \in X; x_i \neq x_j \right\}$$
If $|X| = N_x$, then $|\{d_x\}| = \frac{1}{2}N_x(N_x - 1)$.

The between-class distance (BCD) set, which is the set of distances between any two points from different classes ($X$ and $Y$), can be defined as

$$\{d_{x,y}\} = \left\{ \|x_i - y_j\|_2 | x_i \in X; x_j \in Y \right\}$$
If $|X| = N_x$, $|Y| = N_y$, then $|\{d_{x,y}\}| = N_x N_y$.

Theorem 3.1 shows how the ICD and BCD sets are related to the distributions of the two-class data.

**Theorem 3.1.** *If and only if the two classes $X$ and $Y$ with the same distribution and have sufficient data points, the distributions of the ICD and BCD sets are nearly identical:*

$$\frac{\left|\{d_x = d\}\right|}{\left|\{d_x\}\right|} \cong \frac{\left|\{d_y = d\}\right|}{\left|\{d_y\}\right|} \cong \frac{\left|\{d_{x,y} = d\}\right|}{\left|\{d_{x,y}\}\right|}; \quad \left|\{d_x\}\right|, \left|\{d_y\}\right| \to \infty$$

The proof of Theorem 3.1 is presented in Appendix 7.1. According to this theorem, that the distributions of the ICD and BCD sets are nearly identical indicates that the system has maximum entropy and thus has the lowest separability. The time costs for the ICD and BCD sets increase linearly with the number of dimensions and quadratically with the number of data points. It is much better than computing the system's entropy by dividing the space into many small regions. Our experiments show that the time costs could be greatly reduced using a randomly sampled subset from the entire dataset without significantly affecting the results.

### 3.2 Computation of the DSI

First, the ICD sets of $X$ and $Y$: $\{d_x\}, \{d_y\}$ and the BCD set: $\{d_{x,y}\}$ are computed. To examine the similarity of the distributions of the ICD and BCD sets, we apply the Kolmogorov–Smirnov (KS) test. Although there are other statistical measures to compare two distributions, such as Bhattacharyya distance, Kullback–Leibler divergence, and Jensen–Shannon divergence, most of them require the two sets to have the same number of data points. It is easy to show that the $|\{d_x\}|$, $|\{d_y\}|$ and $|\{d_{x,y}\}|$ cannot be the same. The Wasserstein distance is also a potentially suitable measure, but it is not as sensitive as the KS test as compared in Section 5.1.

The similarities between the ICD and BCD sets are then computed using the KS test: $s_x = KS(\{d_x\}, \{d_{x,y}\})$ and $s_y = KS(\{d_y\}, \{d_{x,y}\})$. Details of the KS similarity calculation are provided in Section 5.1. Since there are two classes, the DSI is the average of the two KS similarities: $DSI(\{X,Y\}) = (s_x + s_y)/2$. In general, for an $n$-class dataset, there are $n$ ICD sets for each class: $\{d_{c_i}\}$  $i = 1, 2, \cdots, n$. For the $i$-th class of data $C_i$, the BCD set is the set of distances between any two points in $C_i$ and $\bar{C}_i$ (other classes, not $C_i$): $\{d_{c_i, \bar{c}_i}\}$. The KS similarity between ICD and BCD set is $s_i = KS(\{d_{c_i}\}, \{d_{c_i, \bar{c}_i}\})$. The DSI of this dataset is $DSI(\{C_i\}) = (\sum s_i)/n$.

# 4 EXPERIMENTS AND RESULTS

We test our proposed DSI measure on two-class synthetic and multi-class real datasets and compare it with other complexity measures from the Extended Complexity Library (ECoL) package [4] in R. Since the DSI is computed using KS tests between the ICD and BCD sets, it ranges from 0 to 1. For separability, a higher DSI value means the dataset is easier to separate *i.e.* it has lower data complexity. Hence, to compare with other complexity measures, we use $(1 - DSI)$. In this paper, higher complexity means lower separability (*i.e.* Separability = 1 − Complexity).

## 4.1 Two-class synthetic data

### 4.1.1 Shape and position

In this section, we present the results of the DSI and the other complexity measures for several typical two-class datasets. Figure 3 displays their plots and histograms of the ICD sets (for Class 1 and Class 2) and the BCD set (between Class 1 and Class 2). Each class consists of 1,000 data points.

TABLE 2 presents the results for these measures. Since they assess complexity, a lower complexity means better separability. The measures shaded in grey are considered to have failed in measuring separability and are not used for subsequent experiments. In particular, the dimensionality and class-imbalance measures do not work with separability in this situation. The feature-based and linearity measures measured the XOR dataset as having more complexity than the Random dataset; since the XOR has much clearer boundaries than Random between the two classes, these measures are inappropriate for measuring separability. N1 and N3 produce the same values for the Spiral, Moon, Circle, and Blob datasets, even though the Spiral dataset is obviously more difficult to separate than the Blob dataset, which is the most separable because a single line can be used to separate the two classes. However, the Cls-Coef and Hubs measures assign the Blob dataset greater complexity than some other cases. In this experiment, N2, N4, T1, LSC, Density, and the proposed measure $(1 - DSI)$ are shown to accurately reflect the separability of these datasets.

### 4.1.2 Decision boundary

In this section, we synthesize a two-class dataset that has different separability levels. The dataset has two clusters, one for each class. The parameter controlling the SD of clusters influences separability, and the baseline is the TD we defined.

We created nine two-class datasets using the *sklearn.datasets.make_blobs* function in Python. Each dataset has 2,000 data points (1,000 per class) and two cluster centers for the two classes, and the SD of the clusters is set between 1 and 9. Along with SD of clusters increasing, distributions of two classes are more and more overlapped and mixed together, thus reducing the separability of the datasets. In this case, separability could be visualized clearly by the complexity of the decision boundary. Figure 5 shows that datasets with a larger cluster SD need more-complex decision boundaries.

In fact, if a classifier model can produce decision boundaries for any complexity, it can achieve 100% training accuracy for all two-class datasets (*i.e.* no two data points from different classes have the same features) but the training steps (*i.e.* epochs) required to reach 100% training accuracy may vary. For a specific model, a more complex decision boundary may need more steps to train. Therefore, the average training accuracy throughout the training process – *i.e.* TD – can indicate the complexity of the decision

TABLE 2

Complexity measures results for the two-class datasets. The measures shaded in grey failed to measure separability.

| Category | Code | Random | Spirals | XOR | Moons | Circles | Blobs |
|---|---|---|---|---|---|---|---|
| Feature-based | F1 | 0.998 | 0.947 | 1.000 | 0.396 | 1.000 | 0.109 |
| | F1v | 0.991 | 0.779 | 0.999 | 0.110 | 1.000 | 0.019 |
| | F2 | 0.996 | 0.719 | 0.996 | 0.151 | 0.329 | 0.006 |
| | F3 | 0.997 | 0.843 | 0.998 | 0.397 | 0.708 | 0.007 |
| | F4 | 0.995 | 0.827 | 0.997 | 0.199 | 0.500 | 0.000 |
| Linearity | L1 | 0.201 | 0.170 | 0.328 | 0.074 | 0.233 | 0.000 |
| | L2 | 0.485 | 0.407 | 0.487 | 0.114 | 0.458 | 0.000 |
| | L3 | 0.469 | 0.399 | 0.486 | 0.055 | 0.454 | 0.000 |
| Neighborhood | N1 | 0.719 | 0.001 | 0.040 | 0.001 | 0.001 | 0.001 |
| | N2 | 0.502 | 0.052 | 0.071 | 0.025 | 0.043 | 0.017 |
| | N3 | 0.500 | 0.000 | 0.019 | 0.000 | 0.000 | 0.000 |
| | N4 | 0.450 | 0.359 | 0.152 | 0.099 | 0.162 | 0.000 |
| | T1 | 0.727 | 0.045 | 0.043 | 0.008 | 0.012 | 0.001 |
| | LSC | 0.999 | 0.976 | 0.934 | 0.840 | 0.914 | 0.526 |
| Network | Density | 0.916 | 0.919 | 0.864 | 0.847 | 0.880 | 0.812 |
| | ClsCoef | 0.352 | 0.343 | 0.267 | 0.225 | 0.253 | 0.332 |
| | Hubs | 0.775 | 0.822 | 0.857 | 0.767 | 0.650 | 0.842 |
| Dimensionality | T2 | 0.001 | 0.001 | 0.001 | 0.001 | 0.001 | 0.001 |
| | T3 | 0.001 | 0.001 | 0.001 | 0.001 | 0.001 | 0.001 |
| | T4 | 1.000 | 1.000 | 1.000 | 1.000 | 1.000 | 1.000 |
| Class imbalance | C1 | 1.000 | 1.000 | 1.000 | 1.000 | 1.000 | 1.000 |
| | C2 | 0.000 | 0.000 | 0.001 | 0.000 | 0.000 | 0.000 |
| **Proposed** | **1 − DSI** | **0.994** | **0.953** | **0.775** | **0.643** | **0.545** | **0.027** |

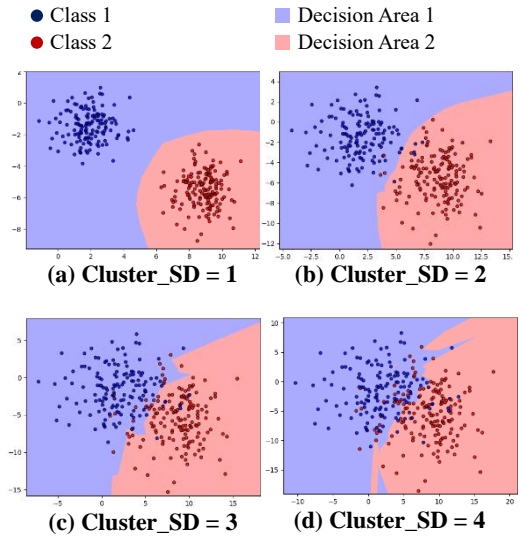

- Class 1
- Class 2
- Decision Area 1
- Decision Area 2

**(a) Cluster_SD = 1**

**(b) Cluster_SD = 2**

**(c) Cluster_SD = 3**

**(d) Cluster_SD = 4**

Figure 5 Two-class datasets with different cluster standard deviation (SD) and trained decision boundaries.

boundary and the separability of the dataset.

We use a very simple fully-connected neural network (FCNN) model to classify these two-class datasets. This FCNN model has three hidden layers, with 16, 32, and 16 neurons with ReLU activation functions in each layer. The classifier was trained on the nine datasets repeatedly, with 1,000 epochs for each training session, to compute the TD for each dataset. Since the training accuracy ranges from 0.5 to 1.0 for two-class classification, to enable a comparison with other measures that range from 0 to 1, we normalize the accuracy by the function:

$$r(x) = (x - 0.5)/0.5$$

$rTD = r(TD)$. The range for rTD is from 0 to 1 with the best score being 1. We also compute N2, N4, T1, LSC, Density, and the proposed measure $(1 - DSI)$ for the nine datasets

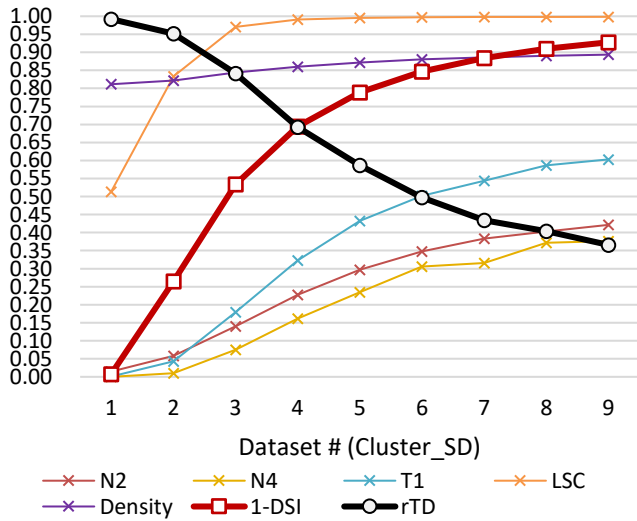

Figure 4 Complexity measures for two-class datasets with different cluster SDs.

and present them together with rTD as a baseline for separability in Figure 4.

As shown in Figure 4, the rTD for datasets with larger cluster SDs is lower. Lower rTD indicates lower separability and higher complexity. N2, N4, T1, and the proposed measure $(1 - DSI)$ reflect the complexity of these datasets well, but the LSC and Density measures do not because they have relatively high values for the linearly separable dataset (#1, see Figure 5a) and increase only slightly for the #5 to #9 datasets. N2, N4, and T1 perform similarly to each other. Compared with them, $(1 - DSI)$ is the most sensitive to the change in separability and has the widest range.

## 4.2   CIFAR-10/100 datasets

The tests in the previous section involve two-dimensional and two-class synthetic data. In this section, we use real images from the CIFAR-10/100 database [16] to examine the separability measures. A simple CNN is used to obtain the baseline TD. This CNN has four convolutional layers, two max-pooling layers, and one dense layer; details of its architecture are presented in Appendix 7.2. The CNN classifier is trained on 50,000 images from the CIFAR-10/100 database. To change the classification performance (*i.e.* the TD), we apply several image pre-processing methods to the images before training the CNN classifier. We hypothesize that these pre-processing methods could change the distribution of the training images and thus alter the separability of the dataset. This change of data separability will affect the classification results for the given CNN in terms of the TD.

Images in the CIFAR-10 dataset are grouped into 10 classes and CIFAR-100 dataset consists of 20 super-classes. Both CIFAR-10 and CIFAR-100 consist of 50,000 training images, each with 3072 features and with each feature being an 8-bit integer. Computing the measures using all 50,000 images would be very time-consuming (including the DSI, most of the measures have a time cost of $O(n^2)$). We test the random selection of a subset of 1/50 training images and find that this strategy does not significantly affect the measures. For example, the DSI for the 50,000 original (*i.e.* without pre-processing) training images from CIFAR-10 is 0.0945, while the DSI for a subset of 1,000 randomly selected images is $0.1043 \pm 0.0049$ – the absolute difference is up to 0.015 (16%) but with an execution speed that is 2,500 times greater. More details about how the size of the subset influences the DSI are presented in Appendix 7.3. In addition, because the same subset is used for all measures, the comparison results are not affected. Therefore, we randomly select 1,000 training images to compute the measures, and this subset still accurately reflects the separability/complexity of the entire dataset.

We use the functions in *PIL.ImageEnhance* with five pre-processing methods applied to original training images from CIFAR-10/100: *Color (factor=2)* and *Sharpness (2)*, *Color (2)*, *Contrast (2)*, *Color (0.1)*, and *Contrast (0.5)*. Including the original images, we use six image datasets to compute the TD and other measures. For the 10-class classification task, the training accuracy ranges from 0.1 to 1.0. The TD is not regularized in this section because it has a range close to [0, 1].

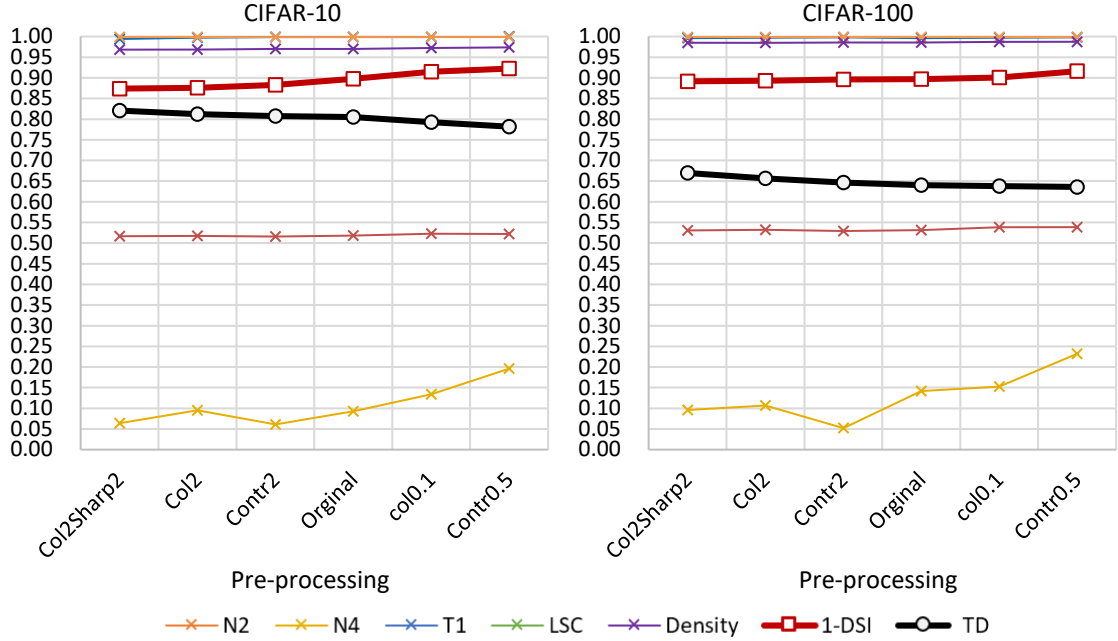

Figure 6 Complexity measures for the CIFAR-10/100 datasets with different pre-processing methods, from left to right: Color (factor=2) and Sharpness (2), Color (2), Contrast (2), Color (0.1), and Contrast (0.5).

Figure 6 shows the results for CIFAR-10 and CIFAR-100. The $x$-axis shows the pre-processing methods applied to the datasets, ordered from left to right by decreasing TD. Since a lower TD indicates lower separability and higher complexity, the values of complexity measures should strictly increase from left to right. LSC, T1 (which almost overlaps with LSC) and Density have high values and remain nearly flat from left to right, while N2 and N4 decrease for the Contrast (2) pre-processing stage. Unlike the other measures, $(1 − DSI)$ monotonically increases from left to right and correctly reflects the complexity of these datasets. These results indicate that image pre-processing is useful for improving CNN performance in image classification because some pre-processing methods improve the separability of the original data.

## 5 DISCUSSION

### 5.1 Kolmogorov–Smirnov tests and other measures

One key step in DSI computation is to examine the similarity of the distributions of the ICD and BCD sets. Theoretically, any measure of the similarity between distributions can be used in this step. We applied the KS test in our study. The result of a two-sample KS test is the maximum distance between two cumulative distribution functions (CDFs):

$$KS(P, Q) = \sup_{x} |P(x) − Q(x)|$$

P and Q are the respective CDFs of the two distributions $p$ and $q$. The $f$-divergence:

$$D_f(P, Q) = \int q(x) f\left(\frac{p(x)}{q(x)}\right) dx$$

cannot be used to compute the DSI because the ICD and BCD have different numbers of values, thus the distributions $p$ and $q$ are in different domains. Measures based on CDFs can solve this problem because CDFs exist in the union domain of $p$ and $q$. Therefore, the Wasserstein distance (W-distance) can be applied as an alternative similarity measure. For two 1-D distributions (*e.g.* ICD and BCD sets), the result of W-distance represents the difference in the area of the two CDFs:

$$W_1(P, Q) = \int |P(x) − Q(x)| dx$$

The DSI uses the KS test rather than the W-distance because we find that normalized W-distance is not as sensitive as the KS test in measuring separability. To illustrate this, we compute the DSI with the two distribution measures for the nine datasets in Section 4.1.2. The two DSIs are then compared with the baseline rTD, which is also used in Section 4.1.2. Figure 7 shows that along with the separability of the datasets decreasing, KS-test has a wider range of decrease than the W-distance. Hence, the KS test is considered a better distribution measure for the DSI in terms of revealing differences in the separability of datasets.

### 5.2 Comparison of distributions

According to Theorem 3.1, the DSI provides another way to verify whether the distributions of two sample sets are identical. As discussed in Section 3.1, if the DSI of sample sets is close to zero, the very low separability means that the two classes of data are scattered and mixed together with nearly the same distribution. The DSI transforms the comparison of $n$-D distributions problem (for two sample sets) to the comparison of 1-D distributions problem (*i.e.* ICD and BCD sets) by computing the distances between samples. For example, in Figure 3(a), samples from Class 1 and 2 come from the same 2-D uniform distribution over $[0, 1)^2$. Consequently, the distributions of their ICD and BCD sets are almost identical and the DSI is about 0.0058. In this case, each class has 1,000 data points. For twice the number of data points, the DSI decreases to about 0.003. When there are more data points of two classes from the same distribution, the DSI will decrease closer to zero, which is the limit of the DSI if the distributions of two samples are identical.

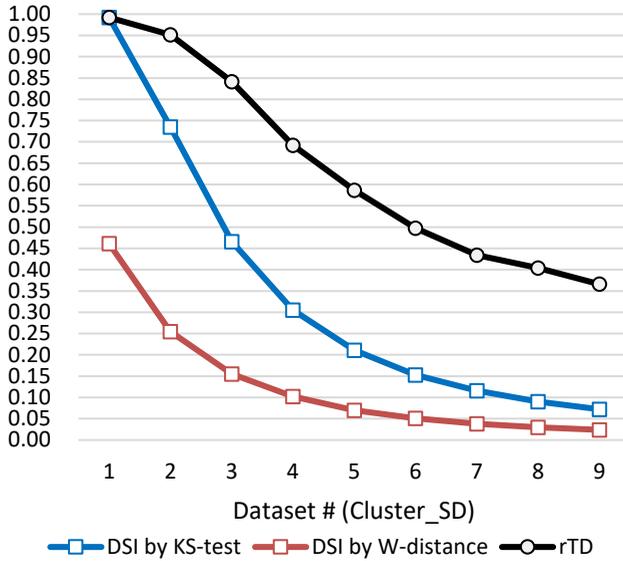

Figure 7 DSI calculated using different distribution measures.

For another example using the CIFAR-10 dataset, we equally divide 5,000 airplane images from this dataset into two subsets: AIR1 and AIR2. We then take a subset of 2,500 automobile images, named AUTO. The "airplane" and "automobile" are class names within CIFAR-10. The DSI of the mixed set: AIR1 (with the label "1") and AIR2 (with the label "2"), is about 0.0045. The DSI of the mixed set: AIR1 (with the label "1") and AUTO (with the label "2") is about 0.1083. Because the images in AIR1 and AIR2 are from the same class and could be considered to have the same distribution, the DSI is closer to zero. In summary, to test if two distributions are identical, we first take labeled data points as many as possible from the two distributions. We then compute the DSI of these data points and see how close the value is to zero. The closer the DSI is to zero, the more likely the two distributions are to be similar.

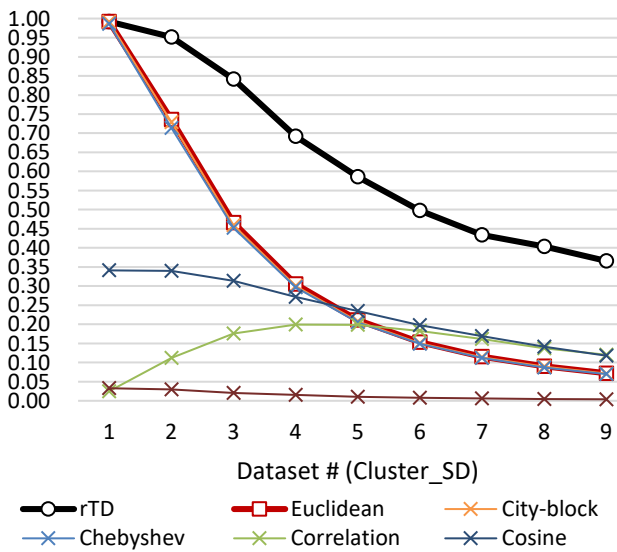

Figure 8 DSI calculated using different distance metrics.

### 5.3 Distance Metrics

Since the DSI assesses the distributions of distances between data points, the distance metric that is used is another important factor. In this study, the DSI uses Euclidean distance. We also test several other commonly used distance metrics: city-block, Chebyshev, correlation, cosine, and Mahalanobis distances. We also computed the DSIs based on these distance metrics using the nine datasets in Section 4.1.2 and the results are compared with the baseline rTD. Figure 8 shows that the performance of Euclidean distance is as similar as those of the city-block and Chebyshev distances. Such result indicates that the Minkowski distance metric ($p$-norm) is suitable for the computation of the DSI.

### 5.4 Future Work and Limitations

The DSI measures the extent of the mixing between data points that belong to different classes. When the DSI is close to zero, it means that the data of different classes have the same distribution. Since DSI can evaluate whether two datasets are from the same distribution, it would be useful to evaluate the generative adversarial networks (GANs) [17]. As with the Fréchet inception distance (FID)[18], measuring how close the distributions of real and GAN-generated images are to each other is an effective approach to assess GAN performance because the goal of GAN training is to generate images that have the same distribution as real images. By examining the similarity of the two distributions, the DSI can detect (or certify) the distribution of a sample set. Several distributions could be assumed (*e.g.* uniform or Gaussian) and a test set is created with an

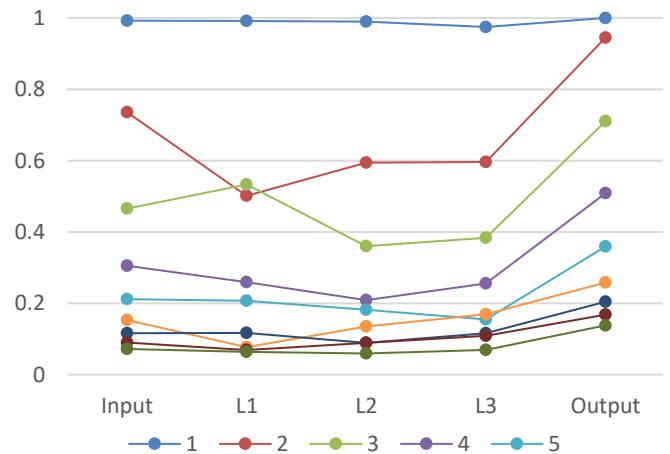

Figure 9 DSIs of input and output data from each layer in the FCNN model for nine datasets. The *x*-axis represents the outputs from layers of the FCNN, and the *y*-axis represents the DSI values of output.

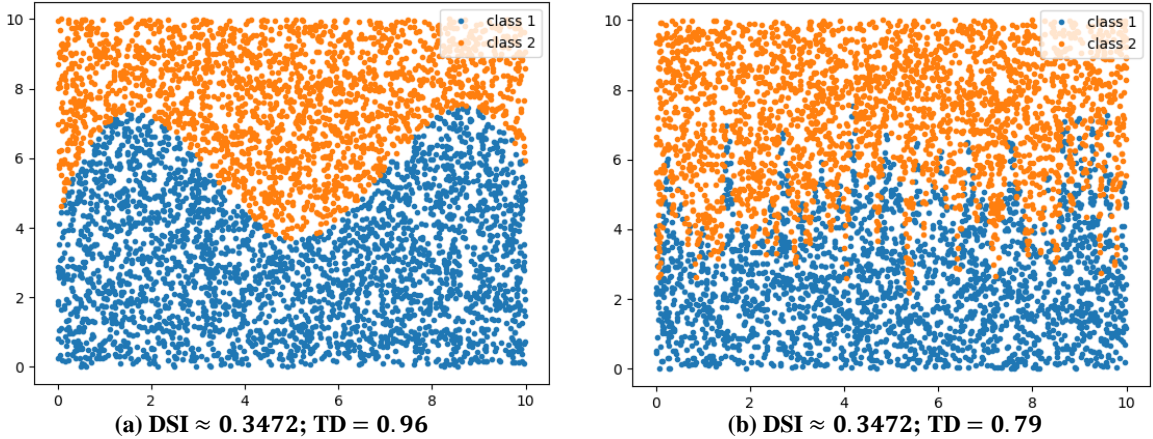

**(a) DSI ≈ 0.3472; TD = 0.96**   **(b) DSI ≈ 0.3472; TD = 0.79**

Figure 10 Two-class datasets with different cluster SDs and trained decision boundaries.

assumptive distribution. The DSI could then be calculated using the test and sample set. The correct assumptive distribution will have a very small DSI (*i.e.* close to 0) value.

The DSI could also help to understand how data separability changes after passing through each layer of a neural network. As an example, we reuse the three-layer FCNN model and nine datasets in Section 4.1.2. An FCNN model is trained using a single dataset. We then input the data into the trained model and record the output from each layer. Finally, we compute the DSI of every output and input data. As shown in Figure 9, for every dataset, that the DSI of final output is always higher than the input indicates the classifier improves the separability of data. Some DSIs of output from hidden layers, however, are even smaller than that of the input data. This phenomenon is non-intuitive because it is assumed that hidden layers improve separability and increase the DSI continuously. A possible reason for this is that the dimensions of data increase in the hidden layers. The dimension of input data is two, and it changes to 16, 32, 16, and 1 for the output because of the number of neurons in hidden layers. In higher dimensional space, data may be coded by fewer features or mapped closer to each other, thus, the separability decreases. In addition, dimensionality can affect data distributions and the measurement of distance, which is known as the curse of dimensionality [19], thus affecting the DSI. These results raise the issue of how to compare separability across different numbers of dimensions.

In addition, the DSI may have other problems that require further research. The DSI can be considered a distance-based embedding method. It extracts one index from data of any *n*-dimension to indicate the separability of that data. By reducing the dimensionality, a significant volume of the original information is lost. The limitations that DSI cannot properly measure the separability of data in some situations must be discovered. For example, when the separability changes, the change of DSI is nonlinear, and the DSI of linearly separable data is usually not 1 (*e.g.* Figure 3f). The DSI may also be influenced by the scale and sparseness of data. The DSI reflects the similarity of distributions of datasets and the datasets having similar distributions are considered difficult to separate.

In fact, our separability is defined by the MEP. In some cases, separability cannot accurately reflect the complexity of the decision boundary. For example, in Figure 10, two datasets have approximately the same DSI but the decision boundary complexity of the right dataset is higher and the TD (from the previously used three-layer FCNN model) is smaller. Therefore, MEP-based separability: 1) cannot be applied to every classification problem such as cases that are simply separable by high-level characteristics, and 2) does not represent the general training difficulty.

## 6 CONCLUSION

Based on MEP, different classes of data that are mixed together with the same distribution constitute the most difficult case to separate using classifiers. Our proposed DSI can indicate whether data belonging to different classes have the same distribution, and thus provides a measure of the separability of datasets. After testing the proposed DSI on synthetic and real datasets and comparing it with several state-of-the-art separability/complexity measures from previous studies, we find that the DSI is a robust and effective separability measure. In addition, DSI may have a wider range of applications in data science, machine learning, and deep learning, such as data distribution estimation, feature selection, understanding the principles of neural networks, and measuring GAN performance.

## 7 APPENDIX

### 7.1 Proof of Theorem 3.1

Consider two classes $X$ and $Y$ that have the same distribution covering the same region and have sufficient data points. Suppose $X$ and $Y$ have $N_x$ and $N_y$ data points, to assume the sampling density ratio is $N_y/N_x = \alpha$. For convenience, we repeat the defininations of ICD and BCD here.

**Intra-Class distance (ICD) set**: *the set of distances between any two points in the same class (X).*

**Between-class distance (BCD) set**: *the set of distances between any two points from different classes (X and Y).*

Before providing the proof of Theorem 3.1, we firstly prove Lemma 7.1, which will be used later.

**Lemma 7.1**. *If and only if two classes $X$ and $Y$ have the same distribution covering region $\Omega$ and $\frac{N_y}{N_x} = \alpha$, for any sub-region $\Delta \subseteq \Omega$, with $X$ and $Y$ having $n_{xi}, n_{yi}$ points, $\frac{n_{yi}}{n_{xi}} = \alpha$ holds.*

**Proof.** Assume the distributions of $X$ and $Y$ are $f(x)$ and $g(y)$. In the union region of $X$ and $Y$, arbitrarily take one tiny cell (region) $\Delta_i$ with $n_{xi} = \Delta_i f(x_i) N_x$, $n_{yi} = \Delta_i g(y_j) N_y$; $x_i = y_j$. Then,

$$\frac{n_{yi}}{n_{xi}} = \frac{\Delta_i g(x_i) N_y}{\Delta_i f(x_i) N_x} = \alpha \frac{g(x_i)}{f(x_i)}$$

Therefore:

$$\alpha \frac{g(x_i)}{f(x_i)} = \alpha \Leftrightarrow \frac{g(x_i)}{f(x_i)} = 1 \Leftrightarrow \forall x_i : g(x_i) = f(x_i) \qquad \square$$

**Proof of Theorem 3.1.**

**Sufficiency:** Within the area, select two tiny non-overlapping cells (regions) $\Delta_i$ and $\Delta_j$ (Figure 11). Since $X$ and $Y$ have the same distribution covering the same region but different densities, the number of points in the two cells $n_{xi}, n_{yi}; n_{xj}, n_{yj}$ fulfills:

$$\frac{n_{yi}}{n_{xi}} = \frac{n_{yj}}{n_{xj}} = \alpha$$

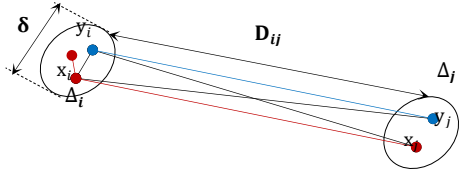

Figure 11 Two non-overlapping small cells.

The scale of cells is $\delta$, the ICDs and BCDs of $X$ and $Y$ data points in cell $\Delta_i$ are approximately $\delta$ because the cell is sufficiently small.

$$d_{x_i} \approx d_{x_i,y_i} \approx \delta; \quad x_i, y_i \in \Delta_i$$

Similarly, the ICDs and BCDs of $X$ and $Y$ data points between cells $\Delta_i$ and $\Delta_j$ are approximately the distance between the two cells $D_{ij}$:

$$d_{x_i,x_j} \approx d_{x_i,y_j} \approx d_{y_i,x_j} \approx D_{ij}; \quad x_i, y_i \in \Delta_i; x_j, y_j \in \Delta_j$$

First, divide the whole distribution region into many non-overlapping cells. Arbitrarily select two cells $\Delta_i$ and $\Delta_j$ to examine the ICD set for $X$ and the BCD set for $X$ and $Y$.

1) **The ICD set** for $X$ has two distances: $\delta$ and $D_{ij}$, and their numbers are:

$$d_{x_i} \approx \delta; \quad x_i \in \Delta_i; \quad \left|\{d_{x_i}\}\right| = \frac{1}{2} n_{xi}(n_{xi} - 1)$$

$$d_{x_i,x_j} \approx D_{ij}; \quad x_i \in \Delta_i; x_j \in \Delta_j; \quad \left|\{d_{x_i,x_j}\}\right| = n_{xi} n_{xj}$$

2) **The BCD set for $X$ and $Y$** also has two distances: $\delta$ and $D_{ij}$, and their numbers are:

$$d_{x_i,y_i} \approx \delta; \quad x_i, y_i \in \Delta_i : \left|\{d_{x_i,y_i}\}\right| = n_{xi} n_{yi}$$

$$d_{x_i,y_j} \approx d_{y_i,x_j} \approx D_{ij}; \quad x_i y_i \in \Delta_i; x_j, y_j \in \Delta_j:$$

$$\left|\{d_{x_i,y_j}\}\right| = n_{xi} n_{yj}; \left|\{d_{y_i,x_j}\}\right| = n_{yi} n_{xj}$$

Therefore, proportions of the number of distances with a value of $D_{ij}$ in the ICD and BCD sets are:

For ICDs:

$$\frac{\left|\{d_{x_i,x_j}\}\right|}{\left|\{d_x\}\right|} = \frac{2 n_{xi} n_{xj}}{N_x(N_x - 1)}$$

For BCDs, considering the density ratio:

$$\frac{\left|\{d_{x_i,y_j}\}\right| + \left|\{d_{y_i,x_j}\}\right|}{\left|\{d_{x,y}\}\right|} = \frac{\alpha n_{xi} n_{xj} + \alpha n_{xi} n_{xj}}{\alpha N_x^2} = \frac{2 n_{xi} n_{xj}}{N_x^2}$$

The ratio of proportions of the number of distances with a value of $D_{ij}$ in the two sets is:

$$\frac{N_x(N_x - 1)}{N_x^2} = 1 - \frac{1}{N_x} \to 1 \quad (N_x \to \infty)$$

This means that the number of proportions of the number of distances with a value of $D_{ij}$ in the two sets is equal. We then examine proportions of the number of distances with a value of $\delta$ in the ICD and BCD sets.

For ICDs:

$$\sum_i \frac{\left|\{d_{x_i}\}\right|}{\left|\{d_x\}\right|} = \frac{\sum_i [n_{xi}(n_{xi} - 1)]}{N_x(N_x - 1)} = \frac{\sum_i (n_{xi}^2 - n_{xi})}{N_x^2 - N_x}$$

$$= \frac{\sum_i (n_{xi}^2) - N_x}{N_x^2 - N_x}$$

For BCDs, considering the density ratio:

$$\sum_i \frac{\left|\{d_{x_i,y_i}\}\right|}{\left|\{d_{x,y}\}\right|} = \frac{\sum_i (n_{xi}^2)}{N_x^2}$$

The ratio of proportions of the number of distances with a value of $\delta$ in the two sets is:

$$\frac{\sum_i (n_{xi}^2)}{N_x^2} \cdot \frac{N_x^2 - N_x}{\sum_i (n_{xi}^2) - N_x}$$

$$= \sum_i \left(\frac{n_{xi}^2}{N_x^2}\right) \cdot \frac{1 - \frac{1}{N_x}}{\sum_i \left(\frac{n_{xi}^2}{N_x^2}\right) - \frac{1}{N_x}} \to 1 \quad (N_x \to \infty)$$

This means that the number of proportions of the number of distances with a value of $\delta$ in the two sets is equal.

In summary, the fact that the proportion of any distance value ($\delta$ or $D_{ij}$) in the ICD set for $X$ and the BCD set for $X$ and $Y$ is equal indicates that the distributions of the ICD and BCD sets are identical and similar to the ICD set for $Y$.

**Necessity:** We prove its contrapositive: if $X$ and $Y$ do not have the same distribution covering the same region, the distributions of the ICD and BCD sets are not identical. We then apply proof by contradiction: suppose that X and Y do not have the same distribution covering the same region, but the distributions of the two sets are identical.

Suppose classes $X$ and $Y$ have the data points $N_x, N_y$: $N_y/N_x = \alpha$. Divide their distribution area into many non-overlapping tiny cells (regions). In the $i$-th cell $\Delta_i$, since distributions of $X$ and $Y$ are different, according to Lemma 7.1, the number of points in the cell $n_{xi}, n_{yi}$ fulfills:

$$\frac{n_{yi}}{n_{xi}} = \alpha_i; \quad \exists \alpha_i \neq \alpha$$

The scale of cells is $\delta$ and the ICDs and BCDs of the $X$ and $Y$ points in cell $\Delta_i$ are approximately $\delta$ because the cell is sufficiently small.

$$d_{x_i} \approx d_{y_i} \approx d_{x_i, y_i} \approx \delta; \quad x_i, y_i \in \Delta_i$$

In the $i$-th cell $\Delta_i$:

1) The ICD of $X$ is $\delta$, with a proportion of:

$$\sum_i \frac{|\{d_{x_i}\}|}{|\{d_x\}|} = \frac{\sum_i [n_{xi}(n_{xi} - 1)]}{N_x(N_x - 1)}$$

$$= \frac{\sum_i (n_{xi}^2 - n_{xi})}{N_x^2 - N_x} = \frac{\sum_i (n_{xi}^2) - N_x}{N_x^2 - N_x}$$

$$(1)$$

2) The ICD of $Y$ is $\delta$, with a proportion of:

$$\sum_i \frac{|\{d_{y_i}\}|}{|\{d_y\}|} = \frac{\sum_i [n_{yi}(n_{yi} - 1)]}{N_y(N_y - 1)} = \frac{\sum_i (n_{yi}^2 - n_{yi})}{N_y^2 - N_y}$$

$$= \frac{\sum_i (n_{yi}^2) - N_y}{N_y^2 - N_y} = \frac{\sum_i (\alpha_i^2 n_{xi}^2) - \alpha N_x}{\alpha^2 N_x^2 - \alpha N_x}$$

$$(2)$$

3) The BCD of $X$ and $Y$ is $\delta$, with a proportion of:

$$\sum_i \frac{|\{d_{x_i, y_i}\}|}{|\{d_{x,y}\}|} = \frac{\sum_i (n_{xi} n_{yi})}{N_x N_y} = \frac{\sum_i (\alpha_i n_{xi}^2)}{\alpha N_x^2}$$

$$(3)$$

For the distributions of the two sets to be identical, the ratio of proportions of the number of distances with a value of $\delta$ in the two sets must be 1, that is $\frac{(3)}{(1)} = \frac{(3)}{(2)} = 1$. Therefore:

$$\frac{\sum_i (\alpha_i n_{xi}^2)}{\alpha N_x^2} \cdot \frac{N_x^2 - N_x}{\sum_i (n_{xi}^2) - N_x}$$

$$= \frac{1}{\alpha N_x^2} \sum_i (\alpha_i n_{xi}^2) \cdot \frac{1 - \frac{1}{N_x}}{\frac{1}{N_x^2} \sum_i (n_{xi}^2) - \frac{1}{N_x}}$$

$$\overset{N_x \to \infty}{=} \frac{1}{\alpha} \cdot \frac{\sum_i (\alpha_i n_{xi}^2)}{\sum_i (n_{xi}^2)} = 1$$

Similarly,

$$\frac{\sum_i (\alpha_i n_{xi}^2)}{\alpha N_x^2} \cdot \frac{\alpha^2 N_x^2 - \alpha N_x}{\sum_i (\alpha_i^2 n_{xi}^2) - \alpha N_x}$$

$$= \frac{\sum_i (\alpha_i n_{xi}^2)}{N_x^2} \cdot \frac{\alpha - \frac{1}{N_x}}{\frac{1}{N_x^2} \sum_i (\alpha_i^2 n_{xi}^2) - \frac{\alpha}{N_x}}$$

$$\overset{N_x \to \infty}{=} \alpha \cdot \frac{\sum_i (\alpha_i n_{xi}^2)}{\sum_i (\alpha_i^2 n_{xi}^2)} = 1$$

To eliminate $\sum_i (\alpha_i n_{xi}^2)$ by considering the two equations, we have:

$$\sum_i (n_{xi}^2) = \frac{\sum_i (\alpha_i^2 n_{xi}^2)}{\alpha^2} \xrightarrow{\rho_i = (\frac{\alpha_i}{\alpha})^2} \sum_i (n_{xi}^2) = \sum_i (\rho_i n_{xi}^2)$$

Since $n_{xi}$ could be any value, to hold the equation, it is necessary that $\rho_i = 1$. Hence:

$$\forall \rho_i = \left(\frac{\alpha_i}{\alpha}\right)^2 = 1 \Rightarrow \forall \alpha_i = \alpha$$

This contradicts $\exists \alpha_i \neq \alpha$. Therefore, the contrapositive proposition has been proved. □

### 7.2 CNN architecture for CIFAR-10/100

TABLE 3
CNN ARCHITECTURE USED IN SECTION 4.2

| Layer | Shape |
|---|---|
| Input: RGB image | 32x32x3 |
| Conv_3-32 + ReLU | 32x32x32 |
| Conv_3-32 + ReLU | 32x32x32 |
| MaxPooling_2 + Dropout (0.25) | 16x16x32 |
| Conv_3-64 + ReLU | 16x16x64 |
| Conv_3-64 + ReLU | 16x16x64 |
| MaxPooling_2 + Dropout (0.25) | 8x8x64 |
| Flatten | 4096 |
| FC_512 + Dropout (0.5) | 512 |
| FC_10 (Cifar-10) | 10 |
| FC_20 (Cifar-100) | 20 |
| Output (softmax): [0, 1] | 10 (Cifar-10)<br>20 (Cifar-100) |

The CNN consists of four convolutional layers, two max-pooling layers, and two fully connected (FC) layers. The activation function for each convolutional layer is the ReLU function, and that for output is softmax function, which maps the output value to a range of [0, 1], with a summation of 1. The notation Conv_3-32 indicates that there are 32 convolutional neurons (units), and the filter size in each unit is 3×3 pixels (height × width) in this layer. MaxPooling_2 denotes a max-pooling layer with a filter of 2×2 pixels window and stride 2. In addition, FC_n represents a FC layer with n units. The dropout layer randomly sets the fraction rate of the input units to 0 for the next layer with every update during training; this layer helps the network to avoid overfitting. Our training optimizer is RMSprop [20] with a learning rate of 1e-4 and a decay of 1e-6, the loss function is categorical cross-entropy, the updating metric is accuracy, the batch size is 32, and the number of total epochs is set at 200.

## 7.3 Subsets and the DSI

CIFAR-10 contains 50,000 images. We randomly choose subsets of 10,000, 5,000, 1,000, 500, and 100 images from CIFAR-10 and compute their DSIs. For each subset, we repeat the random selection of images and computation of the DSI eight times to calculate the SD of DSIs. Computing the DSI for 1,000 images costs about 30 seconds but, for the whole CIFAR-10 training dataset, the DSI calculations cost about 20 hours.

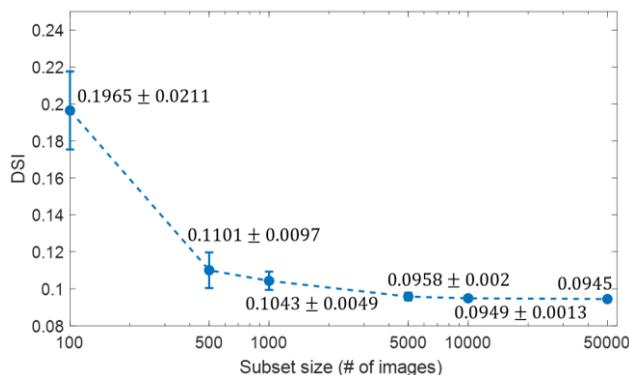

Figure 12 DSI of CIFAR-10 subsets